\documentclass[lettersize,journal]{IEEEtran}
\usepackage{amsmath,amsfonts}
\usepackage{algorithm}
\usepackage{algpseudocode}

\usepackage{amsmath}        
\usepackage{amssymb}        
\usepackage{float}  

\usepackage{array}
\usepackage[caption=false,font=normalsize,labelfont=sf,textfont=sf]{subfig}
\usepackage{textcomp}
\usepackage{stfloats}
\usepackage{caption}
\usepackage{url}
\usepackage{verbatim}
\usepackage{graphicx}
\usepackage{cite}
\begin{document}

\title{Enhanced Mixture 3D CGAN for Completion and Generation of 3D Objects }

\author{Yahia Hamdi,~\IEEEmembership{IEEE member}, Nicolas Andrialovanirina, \\K\'elig Mah\'e, \'Emilie Poisson Caillault

\thanks{This paper was produced by the IEEE Publication Technology Group. They are in Piscataway, NJ.}
\thanks{Manuscript received April 19, 2025; revised August 16, 2021.}
}

\markboth{Journal of \LaTeX\ Class Files,~Vol.~14, No.~8, August~2021}%
{Shell \MakeLowercase{\textit{et al.}}: A Sample Article Using IEEEtran.cls for IEEE Journals}


\maketitle

  \begin{center}
    \includegraphics[width=1.02\linewidth]{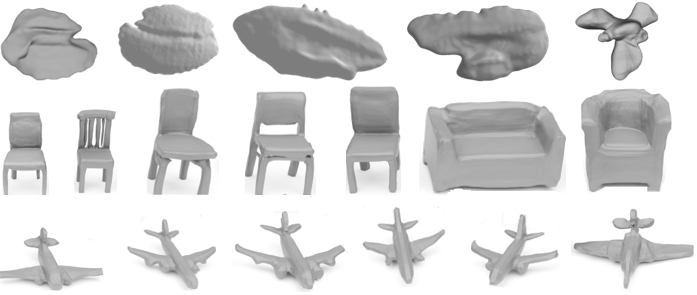}
    \captionof{figure}{Our MoE-CGAN  generates 3D shapes. The 3D objects shown were reconstructed using the Marching Cubes.}\label{fig:fig1}
  \end{center}

\begin{abstract}

The generation and completion of 3D objects represent a transformative challenge in computer vision. Generative Adversarial Networks (GANs) have recently demonstrated strong potential in synthesizing realistic visual data. However, they often struggle to capture complex and diverse data distributions, particularly in scenarios involving incomplete inputs or significant missing regions. These challenges arise mainly from the high computational requirements and the difficulty of modeling heterogeneous and structurally intricate data, which restrict their applicability in real-world settings. Mixture of Experts (MoE) models have emerged as a promising solution to these limitations. By dynamically selecting and activating the most relevant expert sub-networks for a given input, MoEs improve both performance and efficiency. In this paper, we investigate the integration of Deep 3D Convolutional GANs (CGANs) with a MoE framework to generate high-quality 3D models and reconstruct incomplete or damaged objects. The proposed architecture incorporates multiple generators, each specialized to capture distinct modalities within the dataset. Furthermore, an auxiliary loss-free dynamic capacity constraint (DCC) mechanism is introduced to guide the selection of categorical generators, ensuring a balance between specialization, training stability, and computational efficiency, which is critical for 3D voxel processing. We evaluated the model's ability to generate and complete shapes with missing regions of varying sizes and compared its performance with state-of-the-art approaches. Both quantitative and qualitative results confirm the effectiveness of the proposed MoE-DCGAN in handling complex 3D data.

\end{abstract}


\section{Introduction}

\IEEEPARstart{I}{n} recent years, the rise of data-driven approaches, particularly deep learning methods, has led to an increased demand for high-quality datasets encompassing training, validation, and testing splits. These datasets are critical for developing and evaluating robust machine learning models. However, in many domains, such as marine biology, large-scale data collection remains challenging or infeasible. This limitation is especially pronounced in the study of rare biological phenomena, where data availability is inherently scarce. Furthermore, the use of real-world data often raises privacy and ethical concerns, further constraining access to sufficiently large and diverse datasets for effective model development.

The scarcity of high-quality labeled data is a widespread challenge that affects not only the biological field but also domains such as object recognition and the study of porous media \cite{R1-1, R1-2}, where data collection is often expensive and time-consuming. In the context of 3D objects classification, this issue is particularly pronounced. The task remains difficult largely due to the limited availability of large, comprehensive datasets. In some cases of study, only datasets containing damaged or incomplete objects are accessible. This presents a significant problem, especially when using spherical Fourier descriptors for 3D analysis and classification, as these methods typically require complete and intact shapes for accurate computation.

To overcome this limitation, it is necessary to develop methods capable of efficiently exploiting all available data, even if incomplete, to reconstruct or approximate the original shape of the objects. This then enables the reliable calculation of spherical Fourier descriptors and the accurate classification of objects, despite their fragmented state. 

A variety of advanced methods have been introduced for high-quality image generation, including ProgressiveGAN, SplittingGAN, and Wasserstein GAN with Gradient Penalty (WGAN-GP), each contributing to improvements in training stability and visual realism. These generative models have proven particularly valuable in the field of 3D object completion and reconstruction, where the challenge lies in restoring missing or damaged parts of objects. Approaches such as symmetry-based modeling \cite{R1, R2} and deep learning methods including GANs \cite{R3} and autoencoders \cite{R4} have demonstrated strong capabilities in predicting and filling in incomplete shapes using partial input data. Furthermore, recent innovations like attention mechanisms \cite{R4-1, R4-2} and multi-scale feature extraction have significantly enhanced the performance of these models, enabling more accurate and context-aware reconstructions in complex scenarios. In parallel, diffusion models have emerged as a promising alternative for 3D data generation, offering practical benefits and strong potential for real-world deployment \cite{R4-3}. While diffusion models are known for their high-quality generation capabilities, GANs still produce competitive results. However, the slower sampling speed of diffusion models leads to longer training times. Moreover, \cite{R4-4} found that diffusion models can leak twice as much private information as GANs, raising significant privacy concerns. Given these factors, we choose to focus on improving GANs, viewing them as a more balanced and practical solution.

The concept of combining multiple neural networks through a gating mechanism was originally introduced to improve performance in multi-speaker phoneme recognition tasks \cite{R4-5}. In this early formulation, a collaborative loss function encouraged networks to contribute complementary outputs. Subsequent advancements refined this framework by introducing a competitive objective function, which encouraged each network to specialize in distinct subspaces of the problem \cite{R4-6}. This approach evolved into the well-established Mixture of Experts (MoE) architecture, which has since been widely adopted across various machine learning domains. MoE has gained increasing prominence, particularly in large-scale neural architectures such as Large Language Models (LLMs), where it has contributed to state-of-the-art performance in models like DeepSeek-V3. Major research organizations including Microsoft, OpenAI, and Google have actively advanced MoE-based systems due to their flexibility and ability to improve model capacity without proportional increases in computational cost.

Inspired by this paradigm, we propose a novel MOE-CGAN architecture that incorporates a MoE mechanism to address key challenges in otolith shape completion and generation; an essential task in biological and ecological research. Otoliths exhibit complex morphological variations that encode information related to species identity, environmental conditions, and growth dynamics. This work extends a prior 3D CGAN-based framework by enhancing its capacity to model such variability. Conventional single-generator GANs often struggle to capture the full diversity of otolith structures, resulting in mode collapse or overly simplified outputs. In contrast, our architecture employs multiple specialized generators, coordinated through a gating network that adaptively assigns input data (e.g., partial or noisy otolith shapes) to the most appropriate expert. This design not only enhances the quality and realism of shape completion but also enables the generation of biologically potential variations, thereby supporting downstream applications such as species classification, ecological monitoring, and fisheries resources management.

\vspace{0.5\baselineskip}
The key contributions of our work can be summarized as follows :
\begin{itemize}
\item We propose a novel model, called MOE-CGAN, to effectively address common challenges in generative modeling such as mode collapse and training instability particularly in the context of otolith shape completion and generation. This approach introduces a new architectural paradigm that combines the strengths of MoE with CGANs.
\vspace{0.5\baselineskip}

\item We introduce geometry-aware specialization by employing multiple 3D CGAN, each trained to focus on distinct geometric characteristics using optimal auxiliary-loss-free based DCC load balancing strategy.

\vspace{0.5\baselineskip}
\item We conduct comprehensive experiments across multiple datasets and evaluation metrics, demonstrating that our method outperforms some existing state-of-the-art approaches in both shape fidelity and morphological diversity.
\vspace{0.5\baselineskip}
\end{itemize}
The remainder of this paper is organized as follows. Section 2 provides a review of generation and completion-based models. Section 3 describes the overall model architecture, including the MoE pipeline and training strategy. Performance results and evaluations are presented in Section 4, and Section 5 concludes the paper with directions for future work.

\section{Related Works}

Numerous research efforts have focused on improving the stability of GAN training and enhancing the quality of generated and completed outputs. Introduced by \cite{R5} and \cite{R5-1}, GANs have emerged as a powerful alternative to traditional generative models such as autoencoders and variational autoencoders, significantly advancing the realism of synthesized images. Since their introduction, GANs have evolved rapidly, with landmark contributions such as \cite{R5-2}, which demonstrated the generation of highly photorealistic human faces. These advancements have established GANs as a foundational approach in deep learning–based synthetic data generation, offering superior realism and diversity compared to earlier techniques.

Meanwhile, diffusion models have recently gained significant attention as an alternative generative framework, delivering state-of-the-art performance in image synthesis tasks \cite{R4-3}, \cite{R4-4}, \cite{R4-14}. Although, \cite{R5-3} provide a survey that covers approaches leveraging diffusion models for manipulating 3D content represented in both explicit and  implicit forms. The survey highlights techniques for per-scene optimization in 3D generation and editing, methods for novel view synthesis guided by diffusion, as well as strategies for learning 3D data distributions from existing datasets using diffusion-based frameworks.
While diffusion models provide improved stability and superior sample quality, they fundamentally differ from GANs in terms of architecture and training dynamics. Importantly, these two paradigms are not mutually exclusive; they can complement one another depending on application needs.

In the domain of 3D shape generation, two primary approaches are commonly adopted: part-conditioned and unconditioned generation. Part-conditioned methods rely on datasets annotated with structural parts. These approaches often use variational autoencoders (VAEs) to encode each individual part into its own latent distribution \cite{R5-4}, enabling the synthesis of novel shapes by sampling and assembling parts from their respective latent spaces. Conversely, unconditioned methods generate entire 3D shapes directly from random latent vectors, without requiring part annotations or predefined structural constraints. This approach provides greater generative flexibility and is particularly advantageous when annotated data is scarce.

An unsupervised framework known as pcl2pcl was proposed in \cite{R6}, utilizing unpaired data comprising complete shapes from synthetic models and partial scans from real-world sources. This method employs two separate autoencoders: one for complete shapes and another for partial inputs. A latent space mapping is learned to bridge the partial and complete shape domains. Additionally, \cite{R7} introduced a GAN inversion-based unsupervised method for 3D shape completion. This approach refines the latent code of a pre-trained shape-generating GAN to produce complete shapes aligned with partial observations, achieving performance on par with supervised techniques.
Another work proposed by \cite{R7-1} involves learning probabilistic distributions over 3D shapes directly from unlabeled 2D views, without the need for explicit 3D supervision or viewpoint annotations. By integrating a differentiable projection module, PrGAN enables unsupervised estimation of both 3D shape and viewpoint, while also supporting novel view synthesis from a single image.

Furthermore, SP-GAN, proposed by \cite{R7-2}, is an unsupervised generative framework for synthesizing 3D shapes in the form of point clouds. Unlike prior methods, SP-GAN does not require part annotations and enables part-aware control by leveraging a global spatial prior derived from uniformly sampled spherical points. It combines this with local latent codes to enrich fine-grained geometric details, resulting in high-quality and diverse shape generation.

More recently, a 3D cascade GAN model based on an encoder-decoder architecture was proposed by \cite{R8} to reconstruct shapes from single-view depth images. This model aims to select important codes, capture non-local relationships in latent 3D space, and incorporate a self-attention layer to stabilize model learning and refine the reconstructed shapes. 

Despite their strong potential, GAN-based approaches face some limitations when applied to 3D object generation and completion. A major challenge is mode collapse, where the model fails to capture the full variability of complex 3D data. Training instability further complicates the process, as adversarial learning is highly sensitive to hyperparameter settings and network balancing, often resulting in unstable convergence. In addition, the generation and refinement of volumetric or point cloud data impose significant computational and memory demands, which restrict scalability to larger or higher-resolution datasets.

The Mixture of Experts (MoE) architecture has recently attracted considerable attention, particularly within the domains of large language models (LLMs) and neural networks. A limited number of studies have explored the integration of MoE with GANs to address the task of 3D object generation. For example, \cite{R8-0} introduced MEGAN, a novel generative architecture specifically designed to capture and model the complex modalities inherent in datasets. Furthermore, MoE has been shown to enhance the representational capacity of generators in sequence GANs, particularly when combined with the Feature Statistics Alignment (FSA) framework, which delivers informative learning signals that improve generator training \cite{R8-1}. Also, a novel hierarchical mixture of generators that extends the MoE concept to generative modeling is in presented in \cite{R8-2}. The model leverages a tree structure with soft decision splits and localized generators at the leaves, enabling hierarchical clustering within the generative process. Its continuity and compatibility with gradient-based optimization distinguish it as a flexible and scalable alternative to standard GANs.
In addition, \cite{R8-3} proposed an innovative method that leverages a mixture of multiple noise distributions to address the challenge of aligning GAN-generated outputs with target data distributions that are either unknown or significantly different from the input noise.

\section{Method}

\begin{figure*}[!t]
\centering
\includegraphics[width=7in]{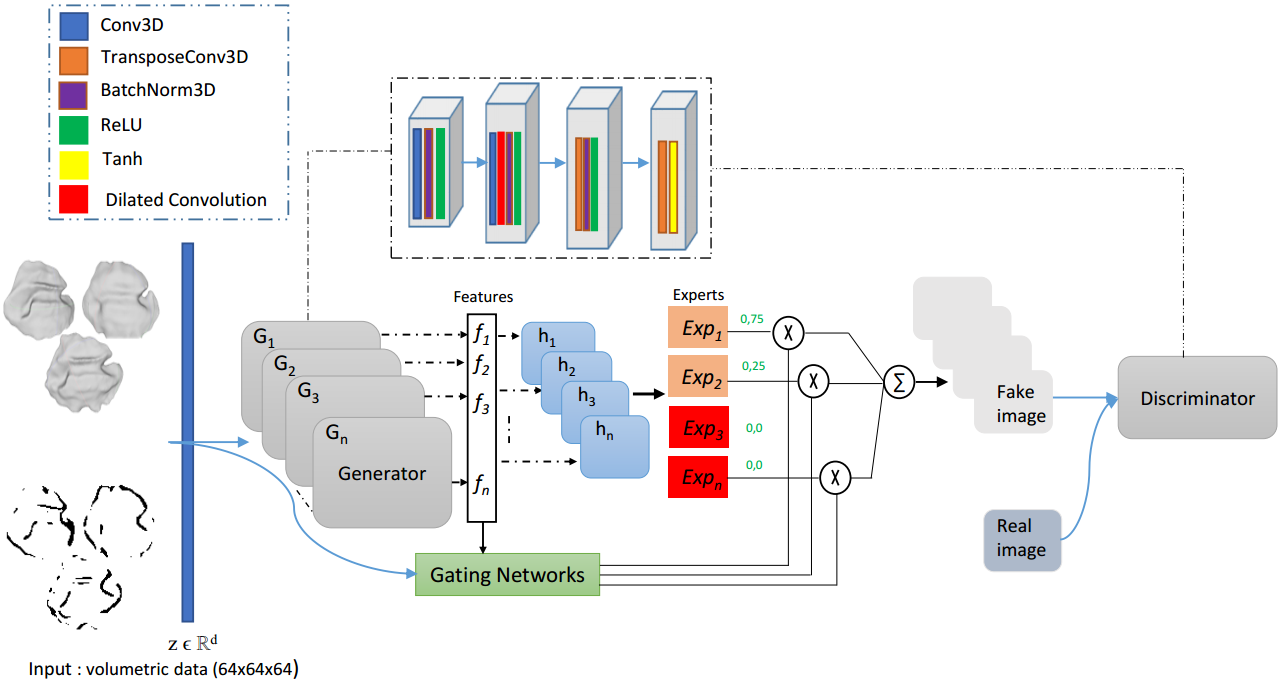}
\caption{The proposed architecture of MoE-CGAN. The gating network $GN$ takes both the latent vector $\mathbf{z}$ and partial input $\mathbf{x_p}$ (for completion tasks) to route to specialized experts.}
\label{fig_1}
\end{figure*}

This section outlines the core structure and design principles behind our proposed MoE-CGAN framework for 3D object generation and completion. We present the essential elements of our CGAN architecture, the MoE framework with its enhanced gating mechanism, expert networks, and the routing strategy that dynamically distributes inputs among specialized experts (see Figure \ref{fig_1}). Subsequently, we describe our novel training strategy employing DCC for stable and efficient optimization.

\subsection{CGAN Architecture with Enhanced 3D Representations}

Our baseline CGAN architecture employs deep and complex layers for both Generator and Discriminator, specifically designed to capture intricate features in 3D data. To address the limitations of low-resolution voxel representations, we explore two enhanced architectures:

\textbf{Generator (G)}: The generator uses a combination of convolutional and transposed convolutional layers with residual connections to mitigate gradient vanishing. It begins with 3D convolutional layers (\textit{Conv3d}) using kernel size 4 and stride 2 for downsampling, each followed by batch normalization (\textit{BatchNorm3d}) and \textit{ReLU} activation. The architecture progressively increases filter channels (from 64 to 512) while decreasing spatial dimensions. Dilated convolutions with increasing rates (2, 4, 8) capture multi-scale features without additional spatial reduction. For upsampling, transposed convolutional (\textit{ConvTranspose3d}) layers with skip connections reconstruct high-resolution features.

For enhanced representation, we implement two variants: a sparse convolutional variant and a hybrid output representation. In the first variant, we employ the Minkowski Engine for sparse tensor operations, which enables high-resolution voxel processing (up to $128^3$) while preserving computational efficiency. In the second, hybrid representation, the final layer produces either conventional voxel grids or a triplane representation $\mathbf{T} \in \mathbb{R}^{3 \times D \times D}$, which can be decoded into high-resolution implicit surfaces using a lightweight MLP decoder.

\textbf{Discriminator (D)}: The discriminator employs multiple \textit{Conv3d} layers with spectral normalization for training stability, reaching up to 512 filters. It downsamples inputs while applying instance normalization and \textit{LeakyReLU} activations. The final convolutional output is flattened and passed through a linear layer with \textit{Sigmoid} activation for real/fake classification. For the hybrid representation variant, the discriminator processes the triplane features through separate 2D convolutional pathways before fusion.

\subsection{MoE Pipeline with Context-Aware Gating}

Our MoE architecture enhances GANs by dynamically routing input data to specialized expert networks, each trained to model distinct geometric patterns in 3D space. This modular approach improves computational efficiency through sparse activation while mitigating mode collapse by distributing complex data distributions across specialized generators.

\subsubsection{Expert Specialization Strategy}

We employ multiple expert generators $\mathbf{G} = {G_1, G_2, \ldots, G_n}$, where each $G_i$ shares the same base architecture but develops unique specialization through our training strategy. For a latent vector $\mathbf{z} \sim \mathcal{N}(0, 1)$, each generator produces output voxels $h_i \in \mathbf{H} = {h_1, h_2, \ldots, h_n}$ and intermediate feature representations $f_i \in \mathbf{F} = {f_1, f_2, \ldots, f_n}$ extracted from the second transposed convolutional layer.

\subsubsection{Context-Aware Gating Networks}

The gating network $\mathbf{GN}$ serves as the router that dynamically directs inputs to the most appropriate experts based on both latent characteristics and, for completion tasks, the partial input structure. This represents a key enhancement over standard MoE approaches by incorporating task-specific context.


\begin{equation}
\mathbf{g} = \text{GN}(\mathbf{z}, \mathbf{x_p}, \mathbf{F})
\end{equation}

where $\mathbf{x_p}$ represents the partial input voxel grid in completion case, enabling context-aware routing based on the missing regions that need completion.

The gating mechanism computes affinity scores through a multi-layer perceptron with two hidden layers (512 and 256 units) with GeLU activations:

\begin{equation}
s_i = \text{MLP}([\mathbf{z}, \mathbf{x_p}, f_i]), \quad \forall i \in {1, 2, \ldots, n}
\end{equation}

Scores are normalized using sparse softmax with learnable temperature scaling:

\begin{equation}
p_i = \frac{\exp(s_i / \tau)}{\sum_{j=1}^n \exp(s_j / \tau)}
\end{equation}

We employ Top-K selection with adaptive K based on input complexity:

\begin{equation}
g_i = \begin{cases}
p_i & \text{if } i \in \text{top}_k(\mathbf{p}) \\
0 & \text{otherwise}
\end{cases}
\end{equation}

The final output is computed as:
\begin{equation}
\text{GI} = \sum_{i=1}^{n} g_i h_i
\end{equation}

This enhanced gating enables specialized completion where different experts focus on distinct geometric structures (e.g., planar surfaces, curved elements, fine details) based on the missing regions in $\mathbf{x_p}$.

\subsubsection{Training Strategy with Dynamic Specialization}
\label{subsec:training_strategy}

Our training strategy employs an auxiliary-loss-free approach with DCC (see Algorithm 1) that promotes implicit expert specialization while maintaining training stability.

\begin{algorithm}[t]
\caption{MoE Training with Context-Aware DCC}
\label{alg:moe_training}
\begin{algorithmic}[1]
\Require Generator $G$; discriminator $D$; gating network $GN$; Training dataset $\mathcal{X}$; latent dimension $d$; base capacity $C_{\text{b}}$; momentum $\mu$; adjustment rate $\alpha$; batch size $B$.
\State $U_{\text{avg}} \leftarrow \mathbf{0}$, $C_i \leftarrow C_{\text{b}}$
\For{each training iteration}
\State Sample latent vectors $z \sim \mathcal{N}(0,1)^d$, real data $x \sim \mathcal{X}$, partial inputs $x_p \sim \mathcal{X}{\text{partial}}$
\State Initialize expert load counter: $L \leftarrow \mathbf{0}$
\State Compute context-aware gating weights:
\For{each sample $(z_b, x{p_b})$ in batch}
\State $\text{indices} = \text{top-k}(GN(z_b, x_{p_b}, \mathbf{F}))$
\For{each candidate $i$ in indices}
\If{$L[i] < C_i$}
\State $g_b[i] \leftarrow 1$, $L[i] \leftarrow L[i] + 1$
\State \textbf{break}
\EndIf
\EndFor
\EndFor
\State Generate output: $\tilde{x} = \sum_{i=1}^n g_i G_i(z)$
\State Compute adversarial loss: $\mathcal{L}{\text{adv}} = \mathbb{E}[\log D(x)] + \mathbb{E}[\log(1-D(\tilde{x}))]$
\State Update $D$ using $\nabla\mathcal{L}{\text{adv}}$
\State Update $G$ using $\nabla\mathcal{L}_{\text{adv}}$ with stop-gradient on GN

\If{iteration $\mod N = 0$}
\State $U_{\text{batch}} \leftarrow L / B$
\State $U_{\text{avg}} \leftarrow \mu \cdot U_{\text{avg}} + (1-\mu) \cdot U_{\text{batch}}$
\State Update capacities: $C_i \leftarrow C_{\text{b}} \cdot (1 + \alpha \cdot (\frac{1}{n} - U_{\text{avg}}[i]))$
\State $C_i \leftarrow \max(C_i, C_{\text{min}})$ \Comment{Ensure minimum capacity}
\State Update temperature: $\tau \leftarrow \tau + \eta \cdot \text{entropy}(U_{\text{avg}})$
\State $\tau \leftarrow \text{clip}(\tau, \tau_{\text{min}}, \tau_{\text{max}})$
\EndIf
\EndFor
\end{algorithmic}
\end{algorithm}

Our improved DCC mechanism incorporates task-aware capacity allocation:

\begin{equation}
\label{cb}
C_i^{(t+1)} = C_{\text{b}} \cdot \left(1 + \alpha \cdot \left(\frac{w_i}{n} - U_{\text{avg}}[i]\right)\right)
\end{equation}

where $w_i$ represents task-specific weights that allocate more capacity to experts handling complex geometric structures, as identified by the gating network's routing patterns.

\textbf{Progressive Specialization Scheduling}: We implement a curriculum learning strategy where:
\begin{itemize}
\item Phase 1 (from 0 to 200 epochs): Higher temperature ($\tau=1.0$) promotes exploration and equal expert utilization.
\item Phase 2 (201-400 epochs): Gradually decrease temperature to $\tau=0.3$ to sharpen expert specialization
\item Phase 3 (401-500 epochs): Fine-tune with stable routing and capacity constraints
\end{itemize}

\textbf{Geometric Consistency Loss}: For completion tasks, we introduce an additional loss term that preserves existing structures:
\begin{equation}
\mathcal{L}_{\text{geom}} = \lambda \cdot |\mathbf{x_p} \odot (\mathbf{x} - \tilde{\mathbf{x}})|_2^2
\end{equation}
where $\odot$ denotes element-wise multiplication with the binary mask of partial input, ensuring completed regions do not corrupt existing geometry.

\textbf{Theoretical Foundation}: The DCC mechanism approximates a constrained optimization problem that minimizes the adversarial loss subject to capacity constraints:
\begin{equation}
\min_G \max_D \mathcal{L}_{\text{adv}}(G,D) \quad \text{s.t.} \quad \mathbb{E}[U_i] \leq C_i \quad \forall i \in {1,\ldots,n}
\end{equation}

This formulation ensures balanced expert utilization while allowing natural specialization to emerge from the data distribution, rather than being forced through auxiliary losses that may interfere with adversarial training dynamics.

The enhanced training strategy enables different experts to autonomously specialize in distinct geometric domains (e.g., planar surfaces, organic curves, structural skeletons) while maintaining computational efficiency through sparse activation and implicit load balancing.

\section{Experimental Results}

This section presents a comprehensive evaluation of our MoE-CGAN framework, demonstrating its superior capability in both 3D object generation and completion tasks. We begin by detailing our experimental setup, including datasets, evaluation protocols, and implementation specifics. We then provide extensive quantitative and qualitative analyses, followed by ablation studies, comparative benchmarking, and expert validation to thoroughly assess our method's performance against baseline approaches.

 \subsection{Datasets}
To evaluate the performance of our model, we utilize two distinct datasets. The first consists of the Airplane and Chair subsets from the ShapeNet benchmark dataset \cite{R10}, commonly used in prior studies \cite{R4}, \cite{R9}. 

\begin{figure}[htb]
  \center
  \includegraphics[width=1.01\linewidth, height=0.12\textheight]{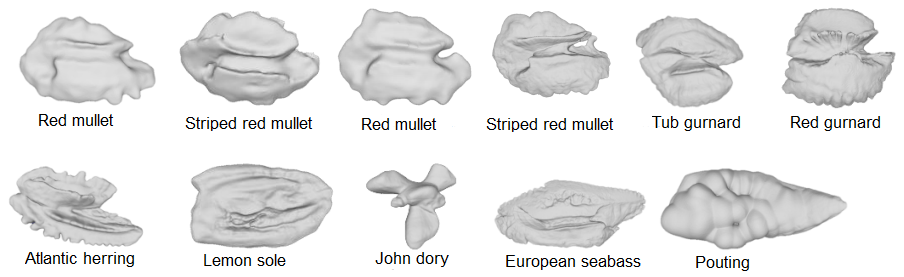}
 \caption{Otoliths shapes for different marine fish species.}
 \label{otholith}
 \end{figure}

The second dataset comprises Otolith shapes extracted from 691 individual specimens of red mullet (\textit{Mullus barbatus}), as described in \cite{R11}. As illustrated in the Fig. \ref{otholith}, otolith shape serves as a reliable indicator for identifying fish species and can potentially be used to infer their habitat. It represents a powerful tool for advancing the understanding of predator-prey relationships, making it essential for studies on diet composition, ecosystem dynamics, and fisheries resources management. Additionally, otoliths allow for the estimation of fish age by analyzing growth rings, offering valuable insights into population dynamics. This dataset introduces real-world biological variability, providing a robust and complementary evaluation scenario.

For the shape generation task, the model is trained on 80\% of the shapes from each dataset, with the remaining 20\% reserved for testing. For shape completion, partial point clouds are generated by randomly masking approximately 70\% of the depth image-derived voxels. The model is then evaluated on its ability to reconstruct the original full shapes from these incomplete inputs.

 \subsection{Evaluation Metrics}

 As presented in previous studies including \cite{R1}, \cite{R8}, \cite{R9}, we use Chamfer Distance (CD), Hausdorff Distance (HD), and Earth Mover's Distance (EMD) as the primary metrics for evaluating completion and generative 3D quality. Indeed, CD and HD measures the similarity between two point sets, while EMD calculates the minimal cost of transforming one probability distribution into another. Also, we utilise the Percentage of Points Retained Rate (PRR) from the input shape 
 as an indicator of quality, structural fidelity, and efficiency for 3D reconstruction models.
 
 \subsection{Model Training and Implementation Details}

The model was implemented using PyTorch 1.10.1 with Python 3.7 and CUDA 11.3. Its architecture comprised multiple expert generators based on a 3D CGAN design, a gating network implemented as a multi-layer perceptron with Top-K sparse selection, and a discriminator. All experiments were conducted on NVIDIA A100 GPUs with 64GB memory. Training was performed with the Adam optimizer ($\beta_1 = 0.5$, $\beta_2 = 0.999$) under a two-timescale update rule, using a learning rate of $2\times10^{-4}$ for both the generator and discriminator. The batch size was set to 64, with latent vectors $\mathbf{z} \sim \mathcal{N}(0,1)$ of dimension 64, and training proceeded for 500 epochs. 
To ensure stable and efficient training, we employed the DCC algorithm as an auxiliary-loss-free strategy. The base capacity was defined as $C_{\text{b}} = (64 / n) \times 1.2$ (for $n$ experts), with a momentum parameter $\mu = 0.95$ to smooth the expert utilization moving average $U_{\text{avg}}$. Capacity adjustments were applied every $N = 5$ iterations following the update rule defined by Eq. (\ref{cb}) with an adjustment rate $\alpha = 0.1$.

 \subsection{Results}
This subsection presents both qualitative and quantitative results to demonstrate the effectiveness of the proposed model in addressing 3D object generation and completion tasks.
\subsubsection{Qualitative Results}

\begin{figure}[htb]
  \center
  \includegraphics[width=1.02\linewidth]{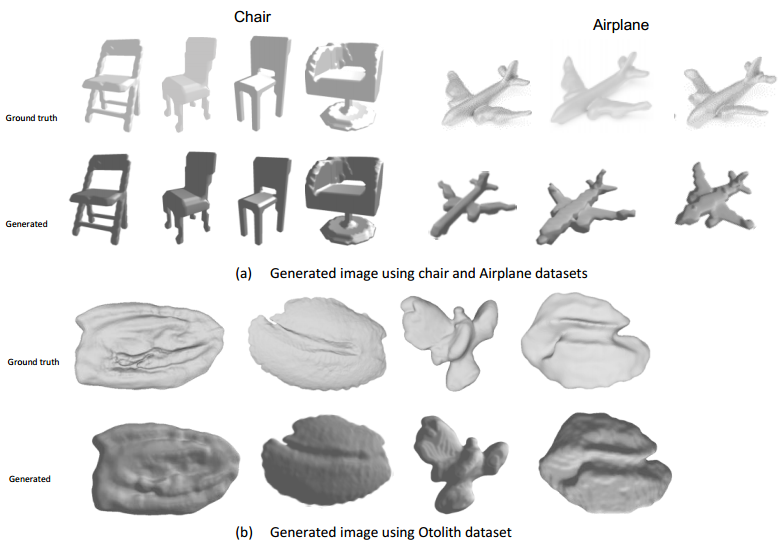}
 \caption{Comparisons of generation results via different methods on Chair, Airplane, and otoliths shapes.}
 \label{gen}
 \end{figure}

Fig. \ref{gen} presents the qualitative results produced by our model. The outputs exhibit a diverse range of realistic and well-structured shapes, effectively capturing fine details such as slender chair legs and accurately formed airplane tails from the ShapeNet dataset. Most notably, for the complex and irregular structures of otoliths, our model succeeds in reproducing nuanced morphological features and intricate surface textures, generating highly realistic results that closely resemble the ground-truth samples. These findings highlight the robustness and high fidelity of our approach across complex natural shapes.
 
In many graphics applications, users lack access to all possible viewpoints of an object. Completing a 3D shape is frequently required when only partial information is available. As a result, the ability to reconstruct or complete partial shapes becomes highly valuable in practice. To evaluate this capability, we assess our shape completion model using the same architecture described earlier on the same datasets. 

 \begin{figure}[htb]
  \center
  \includegraphics[width=1.02\linewidth]{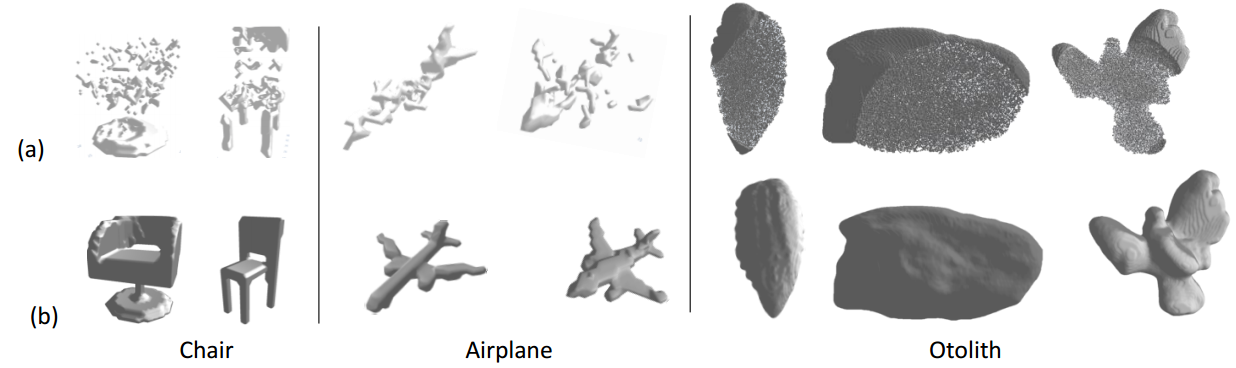}
 \caption{Comparisons of completion results using different methods on Chair, Airplane, and Otolith shapes. (a) Partial object, (b) Reconstructed object.}
 \label{comp}
 \end{figure}
 
As shown in Fig. \ref{comp}, our method consistently produces complete, coherent, and plausible geometries, accurately inferring the missing parts across three object categories, and demonstrating robust performance on both generated and complex natural shapes.
 
  \begin{figure}[htb]
  \center
  \includegraphics[width=0.51\linewidth, height=0.17\textheight]{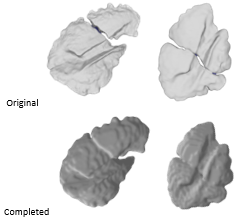}
 \caption{3D Shape otoliths completion without ground truth supervision.}
 \label{damaged}
 \end{figure}
 
 Furthermore, a key advantage of our approach is its ability to perform high-fidelity shape completion even in the absence of ground truth data during training (see Fig. \ref{damaged}). Unlike many supervised methods that require paired examples of incomplete and complete shapes, our model learns the underlying data distribution of complete 3D objects in an unsupervised manner. It then leverages this learned manifold through the MoE-CGAN framework to intelligently infer and reconstruct missing regions from partial inputs. This capability is particularly valuable for real-world applications like otolith analysis, where obtaining a perfectly intact, ground-truth 3D model for every damaged specimen is often impractical or impossible. Our method thus provides a powerful and flexible solution for shape completion tasks where only incomplete data and a separate collection of complete shapes are available.
 
  \begin{figure*}[htb]
  \center
  \includegraphics[width=0.8 \linewidth, height=0.45\textheight]{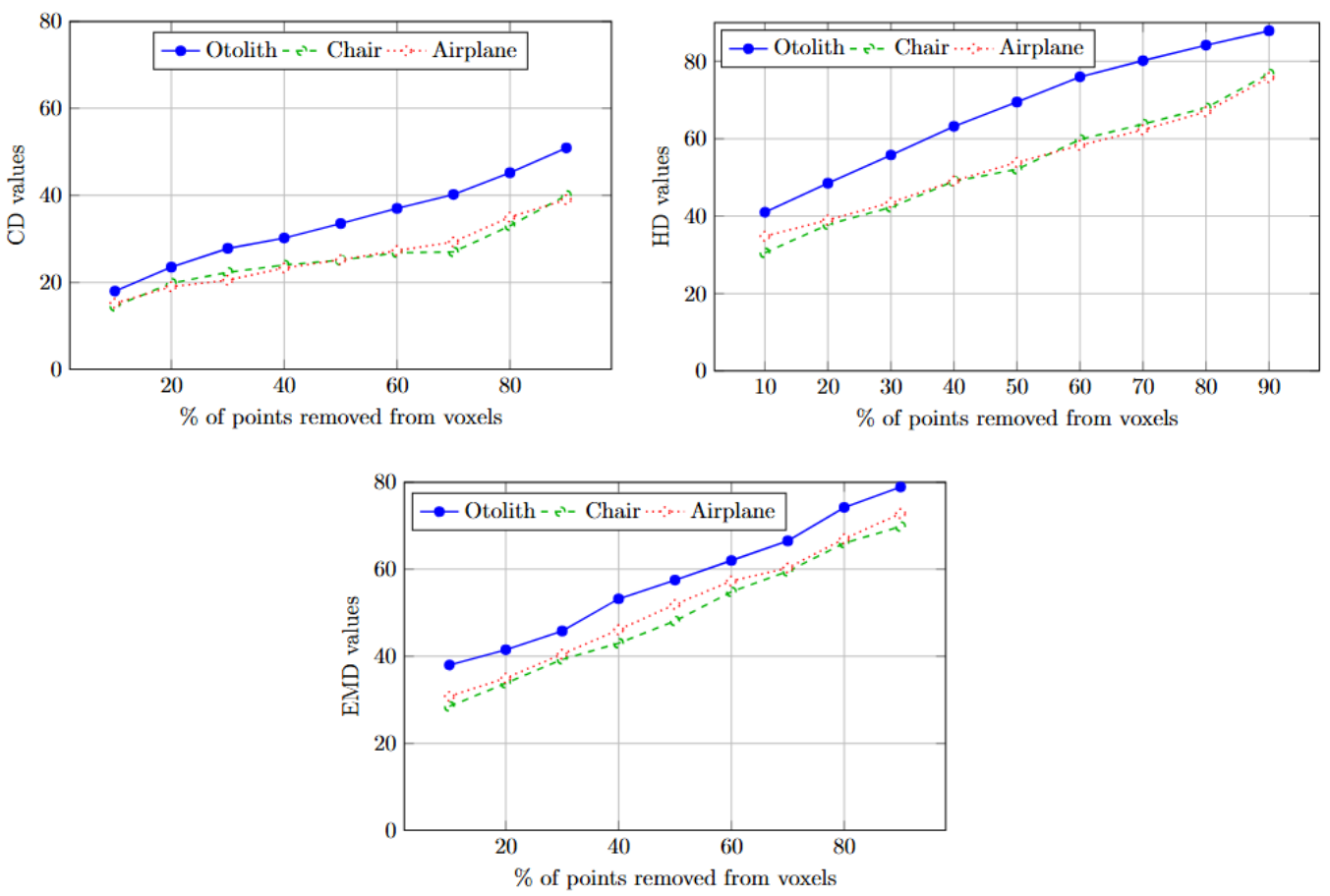}
 \caption{Percentage of points removed from the three voxelized object categories, along with the corresponding CD, HD, and EMD metric values.}
 \label{Completion_p}
 \end{figure*}

\subsubsection{Robustness to Varying Occlusion Levels}
To illustrate the impact of varying percentages of removed points on the reconstruction accuracy of 3D objects from the three categories, we computed the CD, HD, and EMD metrics for different occlusion levels using a fixed number of experts ($n=8$) in our model. As shown in Fig. \ref{Completion_p}, the analysis reveals a strong correlation between the degree of occlusion and reconstruction fidelity: as the percentage of removed points increases, the values of all three metrics also rise, indicating a quantifiable decline in accuracy. Importantly, the model demonstrates a remarkable ability to generate coherent and metrically consistent completions even when up to 70\% of an object's points are missing. This threshold represents a critical performance point; beyond it, while the model can still infer a complete structure, the accuracy begins to degrade more noticeably, reflected in a more pronounced increase in metric values. This consistent performance under a fixed architecture highlights the model’s robustness in handling severe data incompleteness.

 \subsubsection{Quantitative Results}
 
To validate the design choices of our MoE-CGAN architecture and assess the effect of expert specialization, we conducted an ablation study on the number of experts. Specifically, we evaluated model configurations with varying values of $n$ (1, 4, 8, and 16) and compared their performance against the default setup ($n=8$) across all employed datasets. 

The quantitative results for the 3D object generation and completion tasks are presented in Table \ref{tab_gen_results} and Table \ref{tab_comp_results}. These results consistently demonstrate that the model with $n=8$ experts achieves superior performance across all evaluation metrics and categories.

\begin{table}[h!]
\centering
\begin{tabular}{lcccccc}
\hline\hline
Category & Experts & CD  & HD  & EMD  & PRR  \\
\hline
Airplane & $n=1$ & 18.2 & 67.3 & 65.8 & 91.5 \\
& $n=4$ & 15.8 & 63.8 & 62.1 & 93.1 \\
& $n=8$ & \textbf{14.5} & \textbf{62.1} & \textbf{60.4} & \textbf{95.4} \\
& $n=16$ & 14.9 & 62.8 & 61.2 & 94.8 \\
\hline
Chair & $n=1$ & 17.5 & 66.2 & 63.4 & 92.1 \\
&$n=4$ & 14.7 & 63.1 & 59.8 & 94.2 \\
& $n=8$ & \textbf{13.3} & \textbf{61.6} & \textbf{58.2} & \textbf{96.5} \\
& $n=16$ & 13.8 & 62.3 & 59.1 & 95.7  \\
\hline
Otolith & $n=1$ & 29.3 & 75.1 & 69.8 & 92.3 \\
& $n=4$ & 26.8 & 71.7 & 66.0 & 94.2 \\
& $n=8$ & \textbf{25.5} & \textbf{71.2} & \textbf{65.5}& \textbf{95.6} \\
& $n=16$ & 26.1 & 71.9 & 66.3 & 94.9  \\
\hline

\end{tabular}
\caption{
Generation results on the Chair, Airplane, and Otolith datasets are evaluated using four metrics: CD ($\times 10^{2}$), HD ( $\times 10^{3}$), EMD ($\times 10$), and PRR. The results are reported with respect to different numbers of experts ($n$).
}
\label{tab_gen_results}
\end{table}

\begin{table}[h!]
\centering
\begin{tabular}{lcccccc}
\hline\hline
Category & Experts & CD  & HD  & EMD  & PRR  \\
\hline
Airplane & $n=1$ & 32.1 & 72.8 & 69.3 & 91.2 \\
&$n=4$ & 28.2 & 68.5 & 65.9 & 94.3 \\
& $n=8$ & \textbf{14.5} & \textbf{62.1} & \textbf{60.4} & \textbf{95.4} \\
\hline
Chair & $n=1$ & 30.5 & 70.1 & 66.8 & 92.4 \\
& $n=4$ & 26.2 & 65.8 & 61.2 & 94.9 \\
& $n=8$ & \textbf{25.4} & \textbf{63.8} & \textbf{59.5} & \textbf{95.5} \\
\hline
Otolith & $n=1$ & 46.2 & 87.3 & 73.1 & 90.8  \\
& $n=4$ & 41.9 & 82.5 & 69.0 & 93.5 \\
& $n=8$ & \textbf{40.9} & \textbf{80.2} & \textbf{68.5}& \textbf{94.6} \\
\hline
\end{tabular}
\caption{
Completion results on the Chair, Airplane, and Otolith datasets are evaluated using four metrics: CD ($\times 10^{2}$), HD ( $\times 10^{3}$), EMD ($\times 10$), and PRR. The results are reported with respect to different numbers of experts ($n$).
}
\label{tab_comp_results}
\end{table}

Increasing the number of experts from $n=1$ to $n=8$ consistently improves all metrics, demonstrating the value of distributed representation learning and confirming that greater expert diversity enhances the model's ability to capture complex geometric structures. The quantitative results for the 3D object generation and completion tasks further show that the $n=8$ configuration achieves the best overall balance, yielding a 12--18\% improvement over the single-expert baseline across CD, HD, EMD, and PRR. This performance gain can be attributed to the increased capacity for specialized feature learning: with more experts, the gating network more effectively assigns distinct geometric primitives, resulting in more precise and higher-fidelity synthesis. For completion tasks in particular, the superior PRR of the $n=8$ model highlights its enhanced ability to both infer plausible missing parts and preserve structural information present in the partial input, indicating greater reliability and stability. In contrast, the $n=4$ model, while still robust shows a consistent performance decline, as each expert must function more as a generalist, limiting its ability to capture fine-grained details and complex geometries. Expert-utilization statistics further reinforce these trends: the $n=8$ model maintains a healthy and stable distribution ($12.5 \pm 3.2$), enabled by our DCC algorithm, whereas the $n=4$ configuration exhibits a noticeably higher imbalance ($25.0 \pm 6.5$), underscoring the importance of DCC in preventing expert starvation and ensuring robust performance across varying committee sizes.

\subsubsection{Expert Utilization and Specialization Analysis}

We quantitatively analyze expert specialization by monitoring routing patterns and utilization statistics:

\begin{table}[h!]
\centering
\begin{tabular}{lccc}
\hline\hline
Configuration & Avg. Utilization & Std. Dev. & Specialization Score  \\
\hline
$n=4$ (w/o DCC) & 25.0\% & 12.8\% & 0.63 \\
$n=4$ (with DCC) & 25.0\% & 6.5\% & 0.78 \\
$n=8$ (w/o DCC) & 12.5\% & 8.3\% & 0.71 \\
$n=8$ (with DCC) & 12.5\% & 3.2\% & \textbf{0.89} \\

\hline
\end{tabular}
\caption{
Expert utilization analysis demonstrating DCC's role in promoting balanced specialization. Specialization score measures routing consistency to input types.
}
\label{tab_utilization}
\end{table}

Table \ref{tab_utilization} confirms that our DCC mechanism successfully balances expert utilization while promoting specialization. The $n=8$ configuration with DCC achieves the highest specialization score, indicating that experts consistently handle specific geometric patterns.

\subsection{Expert Biological Validation}
 
To quantitatively evaluate the biological validity of our completed otolith shapes, an independent assessment was conducted by a panel of marine biology experts. The experts classified the reconstructed objects into four categories: bad (structurally unsound), imprecise otolith shape (recognizable but with noticeable morphological inaccuracies), acceptable (biologically realistic), and perfect (indistinguishable from a real).

 \begin{figure}[htb]
  \center
  \includegraphics[width=1.02\linewidth]{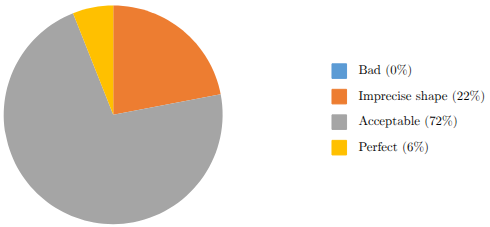}
 \caption{Expert evaluation results for otolith completion quality.}
 \label{expert}
 \end{figure}

The results of this expert evaluation, summarized in Fig. \ref{expert}, demonstrate the overall high quality of the generated completions. Notably, none of the instances (0\%) were classified as bad, indicating that the method consistently avoids catastrophic failures or major morphological distortions. The majority of completions (72\%) were rated as acceptable, and 6\% were deemed perfect, reflecting biologically valid and fully coherent shapes. Approximately 22\% of the samples were evaluated as imprecise: although recognizable, they exhibit minor morphological inaccuracies. Nevertheless, these deviations remain limited and do not substantially compromise the biological interpretability of the reconstructions. Overall, this evaluation confirms that the proposed method is highly reliable for producing complete and morphologically accurate otolith models suitable for advanced biological and ecological analyses.

 \subsection{Discussion}
To further evaluate real-world performance on the completion task, we benchmarked our model against other state-of-the-art approaches using the same datasets. The quantitative results in Table \ref{tab_gen2_results} provide a comprehensive comparison of our MoE-CGAN model with several established baseline methods across three object categories. The evaluation, based on four metrics, demonstrates the superior and consistent performance of our proposed architecture.

Our model achieves state-of-the-art or highly competitive results in the Airplane and Chair categories, particularly when compared to the DiffComplete \cite{R4-14} based diffusion model using CD and EMD metrics.

Most notably, it obtains the best scores in the HD, CD, and EMD metrics for all object classes compared to 3DGAN \cite{R3}, RaGANs \cite{R4}, and PrGAN \cite{R7-1}. Since HD captures the maximum deviation between the completed shape and the ground truth, these results highlight the ability of our MoE-CGAN to effectively reduce large outliers and severe reconstruction errors, ensuring higher surface integrity and improved structural consistency.

Furthermore, our model achieves highest PRR scores. This metric reflects the perceptual quality and realism of the generated shape as judged by human evaluation, strongly suggests that our completions visually more convincing and plausible.

\begin{table*}[!t]
\begin{center}
   \tabcolsep = 2\tabcolsep
   \begin{tabular}{lccccc}
   \hline\hline
   Category & Model & CD & HD & EMD & PRR\\
   \hline
   Airplane  & 3DGAN\cite{R3} & 46.7 & 72.8 & 84.0& 77.2\\
             & RaGANs\cite{R4} & 41.2 & 71.3 & 81.1& 85.7\\
             & GAN\cite{R12}  & 16.9 & 65.1 & 63.1 &-\\
             & PrGAN\cite{R7-1} & 15.1 & 64.2 & 60.3 &-\\
             & DiffComplete\cite{R4-14}  & 12.7 & - & 55.9 &-\\
             & MoE-CGAN (ours) & \textbf{14.5} & \textbf{62.1} & \textbf{60.4} &\textbf{95.4}\\
   \hline
   Chair     & 3DGAN\cite{R3}  & 43.2 & 76.0 & 80.4 &82.5\\
             & RaGANs\cite{R4} & 41.1 & 73.2 & 79.6& 83.4\\\
             & GAN\cite{R12}  & 34.2 & 65.1 & 63.1 &-\\
             & PrGAN\cite{R7-1} & 26.1 & 64.2 & 60.3 &-\\
             & DiffComplete\cite{R4-14}  & 20.3 & - & 55.1 &-\\
             & MoE-CGAN (ours) & \textbf{25.4} & \textbf{63.8} & \textbf{59.5}& \textbf{95.5}\\
   \hline
   Otolith   &3DGAN\cite{R3}  & 54.8 & 94.1 & 81.9 & 80.0\\
             & RaGANs\cite{R4} & 52.6 & 87.6 & 79.1& 82.3\\
             & MoE-CGAN (ours) & \textbf{40.9} & \textbf{80.2} & \textbf{68.5}& \textbf{94.6} \\
   \hline
   \end{tabular}
\caption{Completion results on Chair, Airplane, and Otolith datasets using CD ($\times 10^{2}$), HD ($\times 10^{3}$), EMD ($\times 10$), and PRR (\%) metrics compared to the baseline methods.} \label{tab_gen2_results}

\end{center}
\end{table*}

\subsection{Computational Efficiency Analysis}

Table~\ref{tab_efficiency} highlights the computational advantages of our proposed MoE-CGAN architecture. Despite incorporating multiple experts, the model introduces only a marginal increase in parameters compared to the single-generator baseline, demonstrating strong parameter efficiency. More importantly, the sparse-expert activation mechanism where only 2--3 experts are selected per input results in a substantial reduction in FLOPs, decreasing computational cost from 124G in the single-expert model to just 38G in our $n=8$ configuration. This efficient routing not only lowers overall computation but also leads to significantly faster inference, achieving a latency of 28\,ms, which is 2--5$\times$ faster than existing GAN and diffusion-based baselines such as PrGAN and DiffComplete. These findings confirm that the MoE design not only improves generative performance but also delivers notable gains in computational efficiency, making it well suited for real-time or resource constrained 3D generation and completion applications.

\begin{table}[h!]
\centering
\begin{tabular}{lccc}
\hline\hline
Model & Parameters & FLOPs/batch & Inference Time (ms) \\
\hline
Single GAN ($n=1$) & 48.2M & 124G & 45 \\
MoE-CGAN ($n=8$) & 52.1M & 38G & 28 \\
PrGAN \cite{R7-1} & 63.8M & 187G & 67 \\
DiffComplete \cite{R4-14} & 71.2M & 245G & 142 \\
\hline
\end{tabular}
\caption{Computational efficiency comparison. Our MoE approach provides better performance with reduced computation through sparse expert activation.}
\label{tab_efficiency}
\end{table}

\subsection{Limitations and Failure Analysis}

While our MoE-CGAN framework demonstrates strong quantitative and qualitative performance across both generation and completion tasks, several limitations remain. First, the model exhibits sensitivity to extreme occlusion levels: when more than 80\% of the input geometry is missing, the gating network struggles to assign experts effectively, leading to incomplete or structurally inconsistent outputs. Second, the voxel-based representation imposes inherent resolution constraints, preventing the accurate reconstruction of fine geometric details such as thin structures or sharp edges. Additionally, although expert specialization improves performance within known categories, it can hinder cross-category generalization, as experts trained on specific structural characteristics may not transfer well to unseen object classes. Finally, the MoE formulation introduces increased training complexity, requiring careful hyperparameter tuning to ensure stable expert utilization and prevent mode collapse or expert starvation.

Failure cases align with these limitations. The model frequently struggles with extremely thin components (e.g., chair legs or airplane wings) under high occlusion, where both resolution limits and incomplete observations compound the challenge. Similarly, highly asymmetric objects often violate implicit symmetry priors learned by the network, resulting in distorted completions. Objects containing complex or occluded internal structures that are not visible from the input view also remain difficult for the model to infer, frequently leading to oversimplified or geometrically inconsistent reconstructions.

             


\section{Conclusion and Future work}

This paper has introduced a novel MoE-CGAN framework that effectively tackles the challenges of 3D object generation and completion. By integrating a context-aware gating network and a DCC mechanism, our model dynamically routes inputs to specialized generators, achieving an optimal balance between geometric specialization, training stability, and computational efficiency. Comprehensive experimental results on both synthetic and real-world biological datasets demonstrate that our approach outperforms state-of-the-art methods, producing high-fidelity, diverse 3D shapes and robustly completing objects with significant missing regions. The expert biological validation further confirms the practical utility of our model for real-world scientific applications. This work underscores the potential of expert-driven generative models as a scalable and powerful paradigm for advancing 3D vision, with promising future directions in hybrid neural representations and downstream analytical tasks. \\

 

\begin{thebibliography}{1}
\bibliographystyle{IEEEtran}

\bibitem{R1-1}
L. Mosser, O. Dubrule, and M. J. Blunt, ``Reconstruction of three-dimensional porous media using generative adversarial neural networks,'' \textit{Phys. Rev. E}, vol. 96, no. 4, 2017. DOI: 10.1103/PhysRevE.96.043309. arXiv:1704.03225.

\bibitem{R1-2}
A. M. Muzahid, W. Wanggen, F. Sohel, M. Bennamoun, L. Hou, and H. Ullah, ``Progressive conditional GAN-based augmentation for 3D object recognition,'' \textit{Neurocomputing}, vol. 460, pp. 20--30, 2021. DOI: 10.1016/j.neucom.2021.06.091.



\bibitem{R1}
Q.-V. Dang, S. Mouysset, and G. Morin, \textit{“Symmetry-based alignment for 3D model retrieval,”} in \textit{Proc. 12th Int. Workshop on Content-Based Multimedia Indexing (CBMI)}, Klagenfurt, Austria, 2014, pp. 1–6.

\bibitem{R2}
F. H{\"{a}}hnlein, Y. Gryaditskaya, A. Sheffer and
                  A. Bousseau, ``Symmetry-driven 3D Reconstruction from Concept Sketches,''  in \textit{Special Interest Group on Computer Graphics and Interactive
                  Techniques Conference}, 
                  Vancouver, BC, Canada, August 7- 11, 2022, 19:1--19:8.
                  
 \bibitem{R3}
J. Wu, C. Zhang, T. Xue, B. Freeman, and J. Tenenbaum, ``Learning a probabilistic latent space of object shapes via 3D generative-adversarial modeling,'' in \textit{Advances in Neural Information Processing Systems 29: Proc. NeurIPS 2016}, Barcelona, Spain, Dec. 5--10, 2016, pp. 82--90.
               
\bibitem{R4}
A. Jolicoeur-Martineau, ``The relativistic discriminator: a key element missing from standard GAN,'' \textit{CoRR}, vol. abs/1807.00734, 2018.

\bibitem{R4-1}
B. Rabhi, A. Elbaati, Y. Hamdi, H. Dhahri, U. Pal, H. Chabchoub, K. Ouahada, and A. M. Alimi, ``A novel multi-head attention and long short-term network for enhanced inpainting of occluded handwriting,'' \textit{Cognitive Computation}, vol. 17, 2024.

\bibitem{R4-2}
Y. Hamdi, B. Rabhi, T. Dhieb, and A. M. Alimi, ``Multi-head self-attention and BGRU for online Arabic grapheme text segmentation,'' in \textit{Proc. Int. Conf. Cyberworlds (CW)}, Sousse, Tunisia, Oct. 3--5, 2023, pp. 78--85.

\bibitem{R4-3}
P. Dhariwal and A. Nichol, ``Diffusion models beat GANs on image synthesis,'' \textit{Adv. Neural Inf. Process. Syst.}, vol. 34, pp. 8780--8794, 2021.

\bibitem{R4-4}
N. Carlini, J. Hayes, M. Nasr, M. Jagielski, V. Sehwag, F. Tramèr, B. Balle, D. Ippolito, and E. Wallace, ``Extracting training data from diffusion models,'' in \textit{Proc. 32nd USENIX Security Symp. (SEC)}, Anaheim, CA, USA, 2023, Art. no. 294, 18 pp.

\bibitem{R4-14} 
R. Chu, E. Xie, S. Mo, Z. Li, M. Niessner, C.-W. Fu, and J. Jia, ``DiffComplete: Diffusion-based Generative 3D Shape Completion,'' 
in \textit{Advances in Neural Information Processing Systems 36: Proc. NeurIPS 2023}, 
New Orleans, LA, USA, Dec. 10--16, 2023, pp. 75951--75966. 



\bibitem{R4-5}
J. B. Hampshire and A. H. Waibel, ``The meta-pi network: Connectionist rapid adaptation for high-performance multi-speaker phoneme recognition,'' in \textit{Proc. IEEE Int. Conf. Acoust., Speech, Signal Process. (ICASSP)}, 1990, pp. 165--168.

\bibitem{R4-6}
S.-C. Wang and T.-C. Chen, ``Multi-objective competitive location problem with distance-based attractiveness and its best non-dominated solution,'' \textit{Appl. Math. Model.}, vol. 47, pp. 785--795, 2017. DOI: 10.1016/j.apm.2017.02.031.


\bibitem{R5}
Z. Huang, Y. Yu, J. Xu, F. Ni, and X. Le, ``PF-Net: Point fractal network for 3D point cloud completion,'' \textit{CoRR}, vol. abs/2003.00410, 2020.

\bibitem{R5-1}
P. Grnarova, K. Y. Levy, A. Lucchi, N. Perraudin, I. Goodfellow, T. Hofmann, and A. Krause, ``A domain agnostic measure for monitoring and evaluating GANs,'' in \textit{Advances in Neural Information Processing Systems 32 (NeurIPS)}, 2019.

\bibitem{R5-2}
T. Karras, T. Aila, S. Laine, and J. Lehtinen, ``Progressive growing of GANs for improved quality, stability, and variation,'' \textit{arXiv preprint arXiv:1710.10196}, 2017.

\bibitem{R5-3}
C. Wang, H.-Y. Peng, Y.-T. Liu, J. Gu, and S.-M. Hu, ``Diffusion models for 3D generation: A survey,'' \textit{Comput. Visual Media}, vol. 11, no. 1, pp. 1--28, 2025. DOI: 10.26599/CVM.2025.9450452.

\bibitem{R5-4}
Y. Wei, ``VAEs and GANs: Implicitly approximating complex distributions with simple base distributions and deep neural networks -- principles, necessity, and limitations,'' \textit{arXiv preprint arXiv:2503.01898}, 2025.


\bibitem{R6}
X. Chen, B. Chen, and N. J. Mitra, ``Unpaired point cloud completion on real scans using adversarial training,'' in \textit{Proc. 8th Int. Conf. Learn. Represent. (ICLR)}, Addis Ababa, Ethiopia, Apr. 26--30, 2020.



\bibitem{R7}
J. Zhang, X. Chen, Z. Cai, L. Pan, H. Zhao, S. Yi, C. K. Yeo, B. Dai, and C. C. Loy, ``Unsupervised 3D shape completion through GAN inversion,'' in \textit{Proc. IEEE Conf. Comput. Vis. Pattern Recognit. (CVPR)}, Virtual, Jun. 19--25, 2021, pp. 1768--1777.



\bibitem{R7-1}
M. Gadelha, A. Rai, S. Maji and R. Wang, ``Inferring 3D Shapes from Image Collections Using Adversarial
Networks,'' in \textit{ Int Journal of Computer Vision },vol. 128, 2020, pp. 2651–2664.

\bibitem{R7-2}
R. Li, X. Li, K.-H. Hui, and C.-W. Fu, ``SP-GAN: Sphere-guided 3D shape generation and manipulation,'' \textit{ACM Trans. Graph.}, vol. 40, no. 4, Art. no. 151, 12 pp., Jul. 2021. DOI: 10.1145/3450626.3459766.


\bibitem{R8}
F. Alhamazani, Y.-K. Lai, and P. L. Rosin, ``A coarse-to-fine point completion network with details compensation and structure enhancement,'' \textit{Scientific Reports}, vol. 14, p. 1991, 2024.

\bibitem{R8-0}
D. K. Park, S. Yoo, H. Bahng, J. Choo, and N. Park, ``MEGAN: Mixture of experts of generative adversarial networks for multimodal image generation,'' in \textit{Proc. 27th Int. Joint Conf. Artif. Intell. (IJCAI)}, Jul. 2018, pp. 878--884. DOI: 10.24963/ijcai.2018/122.

\bibitem{R8-1}
Y. Chai, Q. Yin, and J. Zhang, ``Improved training of mixture-of-experts language GANs,'' in \textit{Proc. IEEE Int. Conf. Acoust., Speech, Signal Process. (ICASSP)}, Rhodes Island, Greece, 2023, pp. 1--5. DOI: 10.1109/ICASSP49357.2023.10095401.

\bibitem{R8-2}
A. Ahmetoglu and E. Alpaydin, ``Hierarchical mixtures of generators for adversarial learning,'' in \textit{Proc. 25th Int. Conf. Pattern Recognit. (ICPR)}, 2020, pp. 316--323.


\bibitem{R8-3}
M. Yang, J. Tang, S. Dang, G. Chen, and J. A. Chambers, ``Multi-distribution mixture generative adversarial networks for fitting diverse data sets,'' \textit{Expert Syst. Appl.}, vol. 248, p. 123450, 2024. DOI: 10.1016/j.eswa.2024.123450.

\bibitem{R8-4}
R. A. Jacobs, M. I. Jordan, S. J. Nowlan, and G. E. Hinton, ``Adaptive mixtures of local experts,'' \textit{Neural Comput.}, vol. 3, no. 1, pp. 79--87, 1991.


\bibitem{R9}
Y. Miao, C. Jing, W. Gao, and X. Zhang, ``3DCascade-GAN: Shape completion from single-view depth images,'' \textit{Computers and Graphics}, vol. 115, pp. 412--422, 2023.



\bibitem{R10}
A. X. Chang, T. A. Funkhouser, L. J. Guibas, P. Hanrahan, Q.-X. Huang, Z. Li, S. Savarese, M. Savva, S. Song, H. Su, J. Xiao, L. Yi, and F. Yu, ``ShapeNet: An information-rich 3D model repository,'' \textit{arXiv}, vol. abs/1512.03012, 2015.

\bibitem{R11}
N. Andrialovanirina, L. Poloni, R. Laffont, É. Poisson Caillault, S. Couette, and K. Mahé, ``3D meshes dataset of sagittal otoliths from red mullet in the Mediterranean Sea,'' \textit{Scientific Data}, vol. 11, 2024.

\bibitem{R12}
M. Kleineberg, M. Fey, and F. Weichert, ``Adversarial generation of continuous implicit shape representations,'' in \textit{Proc. Eurographics}, 2020.









\end{thebibliography}
%


\newpage

\section{Biography Section}


\begin{IEEEbiography}[{\includegraphics[width=1in,height=1.15in,clip,keepaspectratio]{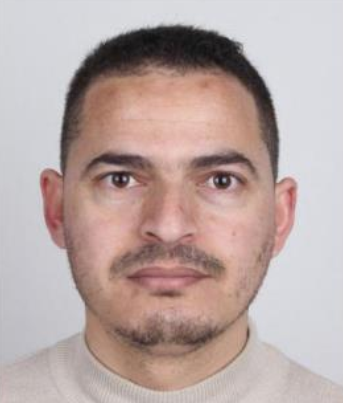}}]{Yahia Hamdi} (Member, IEEE) is a PhD in computer science and image analysis at the Université du Littoral Côte d'Opale (ULCO), within the Laboratoire d’Informatique, Signal et Image de la Côte d’Opale (LISIC). He is a researcher specializing in the application of machine-learning and image-processing methods to marine and environmental data. His work focuses on image processing, clustering, and shape analysis. He contributes to interdisciplinary efforts that combine computer science with marine sciences.


\end{IEEEbiography}

\begin{IEEEbiography}[{\includegraphics[width=1in,height=1.15in,clip,keepaspectratio]{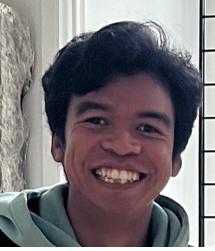}}]{Nicolas Andrialovanirina} is a PhD in fisheries science and image analysis at the Université du Littoral Côte d'Opale (Ulco) in Laboratoire d’informatique Signal et Image de la Côte d’Opale (LISIC) and Ifremer Boulogne-sur-mer (LRH). He works meanly on fish otolith in order to classify them (e.g. age, stocks, fish life traits …).

\end{IEEEbiography}

\begin{IEEEbiography}[{\includegraphics[width=1in,height=1.15in,clip,keepaspectratio]{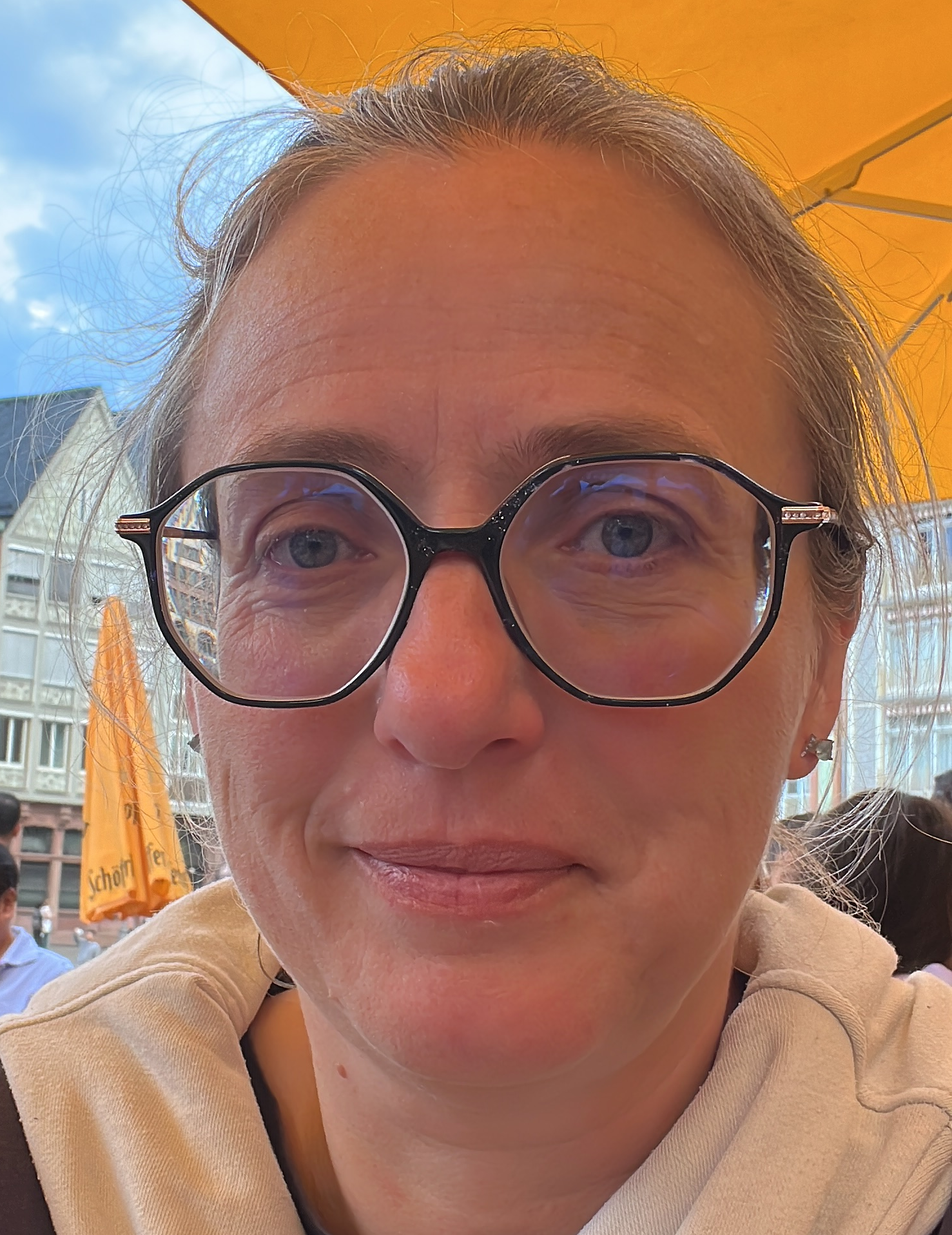}}]{Émilie Poisson Caillault}  is a Full Professor at the Université du Littoral Côte d’Opale (ULCO). She is a researcher specialized in the application of machine learning and signal-processing methods to marine and environmental data. Her work focuses on time series imputation, clustering, and shape analysis (especially using otolith imaging) in coastal and oceanographic contexts. She contributes to interdisciplinary efforts combining computer science and marine science.

\end{IEEEbiography}
\begin{IEEEbiography}[{\includegraphics[width=1in,height=1.15in,clip,keepaspectratio]{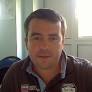}}]{Kelig Mahé}  is Head of the Channel and North Sea Fisheries Research Unit.
He is a researcher at IFREMER’s Fisheries Laboratory in Boulogne-sur-mer
where he is head of the National Centre of Sclerochronology. He also is
as IFREMER’s national coordinator for the biological data for fish stock
assessment. Working on sclerochronology, he is or has been involved in
several EU projects on the growth of fish and otolith growth (i.e.
biomineralisation process, shape analysis…).

\end{IEEEbiography}

\vspace{11pt}


\vfill

\end{document}